\pdfoutput=1

\documentclass[11pt]{article}

\usepackage[final]{acl}

\usepackage{longtable}

\usepackage{times}
\usepackage{latexsym}
\usepackage{amsmath} 
\usepackage{csquotes}
\usepackage{listings}
\usepackage{mathtools}
\usepackage{tabularx}
\usepackage{booktabs}
\usepackage{graphicx}      
\usepackage{subcaption}    
\usepackage{caption}       
\usepackage[T1]{fontenc}

\usepackage[utf8]{inputenc}

\usepackage{microtype}

\usepackage{inconsolata}

\usepackage{graphicx}
\usepackage{amsthm}

\theoremstyle{definition}
\newtheorem{definition}{Definition}[section]

\newcommand{\argsc}{\textsc{AseSum}}
\newcommand{\argscqwen}{${\argsc}_{\text{qwen2.5-7B}}$}
\newcommand{\argscmini}{${\argsc}_{\text{gpt-4o-mini}}$}

\title{Aspect-Based Opinion Summarization with Argumentation Schemes}

\author{Wendi Zhou \and  Ameer Saadat-Yazdi \and Nadin Kökciyan \\
        School of Informatics, \\ University of Edinburgh \\ 
        \{wendi.zhou, ameer.saadat, nadin.kokciyan\}@ed.ac.uk}

\begin{document}
\maketitle
\begin{abstract}
Reviews are valuable resources for customers making purchase decisions in online shopping. However, it is impractical for customers to go over the vast number of reviews and manually conclude the prominent opinions, which prompts the need for automated opinion summarization systems. Previous approaches, either extractive or abstractive, face challenges in automatically producing grounded aspect-centric summaries.
In this paper, we propose a novel summarization system that not only captures predominant opinions from an aspect perspective with supporting evidence, but also adapts to varying domains without relying on a pre-defined set of aspects. Our proposed framework, {\argsc}, summarizes viewpoints relevant to the critical aspects of a product by extracting aspect-centric arguments and measuring their \textit{salience} and \textit{validity}. We conduct experiments on a real-world dataset to demonstrate the superiority of our approach in capturing diverse perspectives of the original reviews compared to new and existing methods.
\end{abstract}

\section{Introduction}



Online reviews are essential resources for customers to make purchase decisions, as they more authentically reflect the performance of some products or services \cite{9183355, amplayo-etal-2021-aspect}. It is very impractical for users to go over most reviews one by one and conclude the prominent opinions discussed themselves. 
Ideally, users should have access to automated opinion summaries to make informed decisions. 

Automatic opinion summarization offers a solution by aggregating all reviews into a concise, easy-to-read summary. Previous methods concerning opinion summarization can be mainly classified as either extractive or abstractive. We see drawbacks with both approaches. Extractive methods select the representative sentences from the input to generate the summary. Although attributable and scalable, they
could encounter issues in generating concise and coherent summaries. 
On the other hand, abstractive methods using neural models to generate fluent and novel summaries may lead to hallucinated content that is challenging to detect without any supporting evidence. 
\citet{hosking2023attributablescalableopinionsummarization} implement a hybrid summarization system, \textbf{\sc{hercules}}, that produces summaries reflecting the general feedback of all reviewers while abstracting away too many details. Although being abstractive and attributable, their summaries are too general for users interested in certain aspects of the entity.

We argue that an ideal summary should reflect the main opinion expressed in the reviews, be attributable with grounding evidence and include critical aspect information that is essential to assist customers 
while making their purchase decisions. 
Many attempts have been made to incorporate aspect information inside the final summary \cite{amplayo-etal-2021-aspect, tang-etal-2024-prompted, li2025aspectawaredecompositionopinionsummarization}; however, they either rely on the manually pre-defined aspects or they lose track of the supporting evidence with a fully automated pipeline using large language models (LLMs).

  To address these existing limitations, we propose an aspect-centric review summarization framework, {\argsc}, to produce high-quality opinion summaries for products. With the help of argumentation schemes and LLMs, {\argsc} extracts aspect-centric arguments, where the claim is the user's sentiment towards certain aspects, and the premise is the supporting evidence mentioned by the users in the reviews.
 This makes the summarization model more generalisable than previous systems as it can easily adapt to new domains, does not require a pre-existing taxonomy of new aspects and can scale up with the number of reviews. By clustering claims supported by similar pieces of evidence, we define a metric to measure the salience and validity of an argument. This metric is used to rank the arguments having the critical aspects information from which we generate our final summaries. In this paper, our main contributions can be summarized as follows:


\begin{itemize}

\item[-] We develop a new automated method that can iteratively induce the aspect taxonomy within the product reviews; 
\item[-]  We introduce a new domain-independent argumentation scheme for aspect-centric argument extraction from customer reviews;
\item[-] We propose a novel hybrid review summarization framework ({\argsc})\footnote{All the code is available online at: \url{https://git.ecdf.ed.ac.uk/s2236454/asesum}} to generate textual summaries. Our model outperforms the current state-of-the-art by 6\% on average on a real-world benchmark dataset.
\end{itemize}

Our paper is organised as follows. We discuss related work on summarization and argumentation in NLP in Section~\ref{sec:rw}. We introduce our review summarization framework~({\argsc}) in Section~\ref{sec:framework}, and Section~\ref{sec:exp} explains our experimental setup before we compare our approach to other models. In Section~\ref{sec:eval}, we show that {\argsc} outperforms these models, not only in terms of the amount of semantic information captured by our summaries but also in the diversity of viewpoints presented. We conclude our paper with a discussion in Section~\ref{sec:conc}.

\section{Related work}\label{sec:rw}

Earlier work on opinion summarization, or review aggregation, is either purely extractive \cite{mihalcea-tarau-2004-textrank, rossiello-etal-2017-centroid, Alguliyev2019COSUM, Belwal2021GraphSummarization} or abstractive \cite{ganesan-etal-2010-opinosis, brazinskas-etal-2020-unsupervised}. 
However, both types of methods have their own shortcomings: extractive methods tend to introduce unnecessary details and struggle to cover all topics in multi-topic inputs, while abstractive methods are limited by the input length of neural models or language models and may generate hallucinated content. \citet{hosking2023attributablescalableopinionsummarization} introduce a hybrid approach, where they encode the review sentences as a hierarchy of paths and then decode the most frequent path in the hierarchy structure as the final summary. Though being unsupervised and attributable, their hierarchy encoder is domain-dependent, thus limiting its generalisability.
Their approach mainly focuses on the general summary generation, neglecting aspect-relevant information.

\citet{10.1162/tacl_a_00366} propose an extractive method that generates aspect-specific summaries using the quantized transformer. Similarly, \citet{amplayo-etal-2021-aspect} develop an abstractive method where they fine-tune a Pre-trained Language Model with aspect controllers for abstractive summaries generation. However, these methods extract aspects either directly from the sentence or with the assistance of humans. Recently, LLMs have demonstrated great performance across a wide range of natural language understanding tasks. Leveraging this, \citet{tang-etal-2024-prompted} propose a fully automated aspect extraction approach through few-shot prompting. They successfully extract aspects together with users' sentiment towards that aspect from reviews; then, after clustering the <aspect, sentiment> pairs, they re-prompt LLMs to generate the aspect-specific keypoints as the final summaries. In this way, they achieve flexible aspect-centric summaries generation at scale, but this iterative prompting pipeline makes their summaries harder to validate without grounding evidence. In contrast, {\argsc} framework preserves the same versatility while providing the grounding evidence by considering argumentative structure. In \citet{li2025aspectawaredecompositionopinionsummarization}, they propose a more explainable and grounded summarization pipeline through prompting LLMs, which separates the
tasks of aspect identification, opinion consolidation, and meta-review synthesis. However, their system requires a set of manually pre-defined aspects, while our system incorporates a flexible aspect induce approach. 

Argumentation schemes have been widely studied in computational argumentation, aiming to model, extract, and generate human-like arguments. A foundational basis for this theory comes from Walton, where he defines structured patterns of common reasoning used in everyday discourse \cite{Walton2008Argumentation}. Each scheme is provided with a template for constructing arguments and critical questions for evaluating their validity. More recent approaches incorporate Walton’s schemes into neural models to guide argument structure prediction and improve the interpretability of human conversations \cite{779cf43f2f974b9f99029d49367b097c}. 

In the context of product reviews, \citet{wyner_adam_semi-automated_2012} introduce a scheme for product reviews based on customer values for semi-automated review analysis. Similarly, \citet{reed_applying_2024} use the \textit{Position to Know} scheme and associated critical questions to evaluate the quality of reviews. We find both these approaches limited in that they ignore the particular features (aspects) of a product that users are discussing, making the analysis too coarse-grained and the evaluation criteria difficult to apply automatically. In contrast, we base our method on a scheme based on \textit{Argument from Characteristic Sign} which we make specific to our aspect/sentiment framework. Our approach also does not depend on critical questions and instead uses an evidence consistency measure to identify the most salient evidence to provide to a user.

\begin{figure*}[t]
\begin{subfigure}{\textwidth}
    \centering
    \includegraphics[scale=0.53]{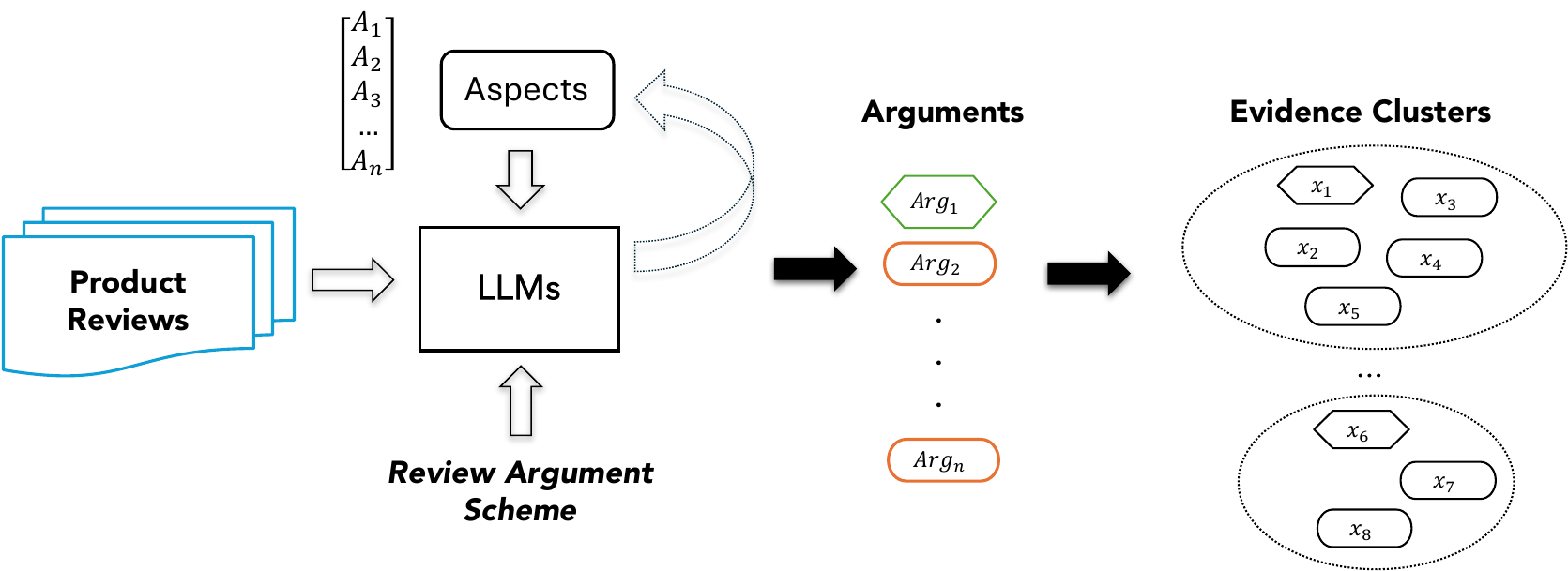}
    \caption{A demonstration of the argument extraction, where we feed the product reviews, the defined Review Argument Scheme together with the set of aspects into LLMs to generate aspect-centric arguments~(Definition~\ref{def:arg}). The aspect set is initiated by prompting LLMs, and is updated during the argument extraction. The arguments are then clustered based on their evidence.}
    \label{fig:stage1}
\end{subfigure}
\begin{subfigure}{\textwidth}
\vspace{2em}

    \centering
    \includegraphics[scale=0.44]{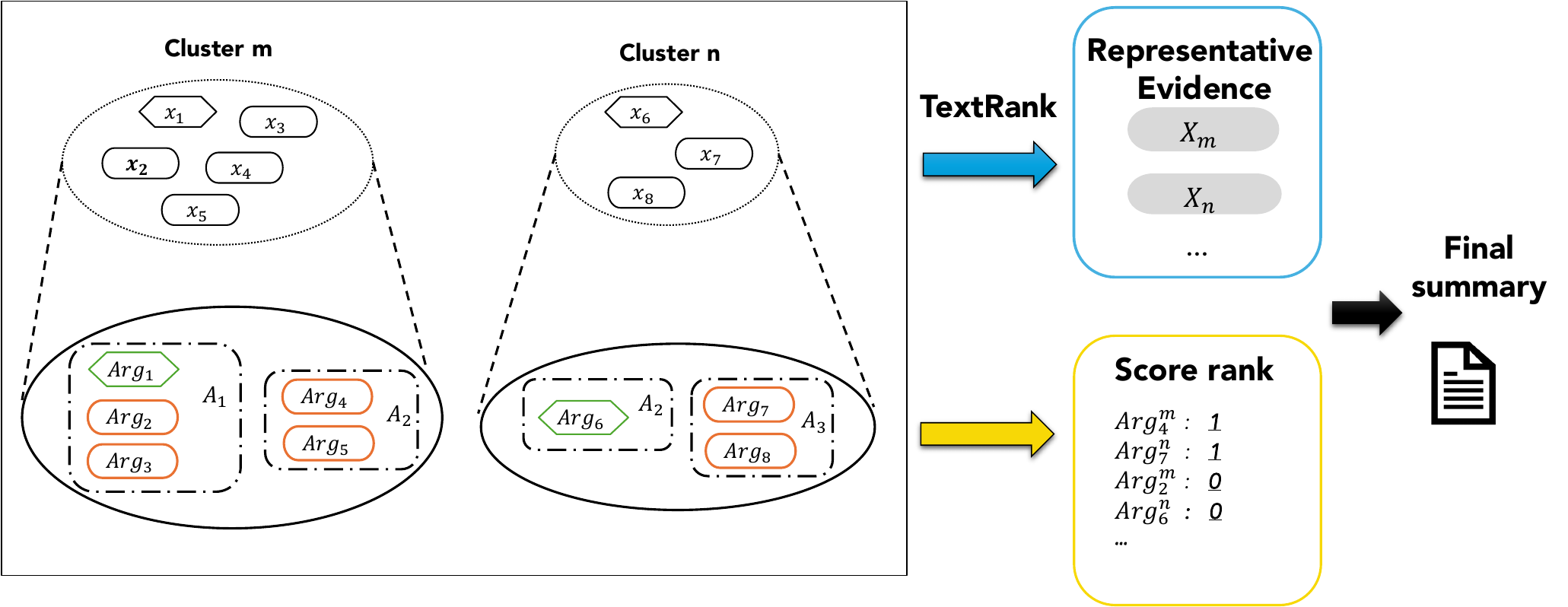}
    \caption{A diagrammatic representation of our methodology starting from clustered arguments. For each clustered argument, {\argsc} selects the representative piece of evidence $X_k$ using the TextRank algorithm. This representative evidence is then used to replace the original evidence of each argument in the cluster. Meanwhile, the system builds the relations among arguments within the same cluster based on their aspects, which are used to measure the salience and validity of the argument as defined in Equation~\ref{eq:val}. Finally, our system selects \textbf{N} unique pieces of evidence from the top-ranked arguments to generate the summary. }
    \label{fig:stage2}
    \end{subfigure}
    \caption{The Proposed {\argsc} Framework} \label{img:argsum}
\end{figure*}

\section{{\argsc} Framework}
\label{sec:framework}


In this section, we introduce an aspect-centric review summarization framework, {\argsc}. The framework has three stages: (i) aspect-centric argument extraction with a new argumentation scheme, \textit{Review Argument Scheme}, (ii) argument clustering and evidence unification, and (iii) argument scoring guided by aspect-centric argument relations.

\subsection{Argument extraction}
\label{sec:3.1:aspect}
Inspired by the argumentation schemes defined by \citet{Walton2008Argumentation}, we propose a novel argumentation scheme for product reviews as shown in Table~\ref{tab:review_scheme}. The Review Argument Scheme~(RAS) consists of three variables: the aspect~(\textbf{A}), the sentiment~(\textbf{S}) and the evidence~(\textbf{X}). In our framework, \textbf{S} takes values from \{\textit{good, bad}\}.

In {\argsc}, each argument is defined as an instantiation of RAS, Definition~\ref{def:arg} provides a formal definition of an argument. Note that $Arg_{i}$ is used to define the \textit{i}th argument.



\begin{definition}[Argument]\label{def:arg}
    $Arg$ denotes a tuple $\langle{a, s, x}\rangle$, where $a$, $s$, $x$ represent the aspect, sentiment and supporting evidence respectively, as they appear in the instantiated argument 
    scheme $Arg$. 
\end{definition}

\begin{table}[tb]
\centering
\begin{tabular}{p{0.45\textwidth}}
\toprule
\textbf{Review Argument Scheme~(RAS)}\\
\midrule
\textit{Claim}: \textbf{A} of this product is \textbf{S}\\
\textit{Major Premise}: \textbf{X} is a sign that \textbf{A} is \textbf{S} \\
\textit{Minor Premise}: The user observes \textbf{X} about \textbf{A} \\
\bottomrule

\end{tabular}
\caption{Proposed argument scheme where \textbf{A}, \textbf{S}, \textbf{X} represent the aspect, sentiment and evidence respectively.}
\label{tab:review_scheme}
\end{table}

In order to instantiate RAS, we benefit from LLMs to fill in the scheme variables and generate arguments with provided user reviews. To avoid LLMs generating diverse aspect representations, we first prompt LLMs to initiate the \textit{critical aspects} of the product given the product category information. The critical aspects represent the key evaluation factors of the product, which may greatly influence customers' purchase decisions. Then we feed them as options into the prompt to guide LLM on performing aspect-centric argument extraction~(Figure~\ref{fig:stage1}). However, for a small subset of reviews, LLMs fail to generate any valid arguments. As this affects only around 3\% of the reviews per domain, it does not have a big influence on the final results.
After obtaining all the arguments extracted by LLMs, we further unify the representations of aspects by clustering them and representing each cluster with a symbol ($A_1, A_2...A_n$). We will provide implementation details in Section~\ref{sec:impl}.

\subsection{Evidence-based Clustering}
\label{sec:3.2:cluster}

Since the evidence for each argument is extracted or slightly summarized by LLMs, it is highly unlikely they will have the same evidence even for arguments sharing the same aspect and sentiment. 
And so, we cluster arguments sharing semantically close evidence and then select the most representative evidence $X_{k}$ for each cluster (Figure \ref{fig:stage2}). 
We assume the most representative evidence of the cluster is the one that entails the majority of the evidence in the cluster. To achieve this, we 
 build a text graph where the vertices are the semantic embeddings of each sentence, and the edge weights are calculated by the cosine similarity between each node pair. Then we iterate the graph-based ranking algorithm derived from Google’s PageRank \cite{page1999pagerank} as described in TextRank \cite{mihalcea-tarau-2004-textrank} until convergence. Finally, we select the vertex with the highest score as the most representative evidence of this cluster.

For each argument $Arg_{i}$ in a cluster $c$, we then substitute its original evidence ($Arg_{i}.x$) with the representative evidence ($X_{c}$) and we rewrite the argument as $Arg_{i}^{c}$. In other words, the evidence of each argument in a cluster is replaced with the most representative evidence. This methodology is depicted in Figure~\ref{fig:stage2}. For example, in \textit{Cluster~m}, we see five arguments, where each of them is supported with its unique evidence. $X_{m}$ would be the representative evidence for \textit{Cluster~m}. If $Arg_{1}$ is represented as $\langle a_1, s_1,x_1 \rangle$, this argument would be rewritten as $Arg_{1}^{m}$=$\langle a_1, s_1,X_m \rangle$. All other arguments could be rewritten similarly. 



After unifying similar evidence for every argument, we calculate a score for each argument based on its popularity and its validity in supporting or opposing a claim related to an aspect.

\subsection{Aspect-centric Ranking}
\label{sec:3.3:score}
To quantitatively assess the salience and validity of an argument, we make use of its support and contradiction relations to other arguments in the same cluster (Section \ref{sec:3.2:cluster}). Firstly, we provide formal definitions of the relations between arguments in Definitions~\ref{def:support} and~\ref{def:contr}. 

\begin{definition}[Aspect-centric Support]\label{def:support}
    A support relation between two arguments in the same cluster, $Arg_{i}$ and $Arg_{j}$, exists if and only if both arguments have the same aspect (i.e., $Arg_{i}.a=Arg_{j}.a$) and sentiment (i.e., $Arg_{i}.s=Arg_{j}.s$).
\end{definition}

\begin{definition}[Aspect-centric Contradiction]\label{def:contr}
    A contradiction relation between two arguments in the same cluster, $Arg_{i}$ and $Arg_{j}$, exists if and only if both arguments have the same aspect (i.e., $Arg_{i}.a=Arg_{j}.a$) and different sentiment (i.e., $Arg_{i}.s  \neq Arg_{j}.s$).
\end{definition}




Intuitively, we consider an argument to be strengthened when a similar evidence supports the same claim from another argument, and an argument to be weakened if a similar evidence is used to support the opposite claim from another argument.
For example, for a pair of shoes, a piece of evidence could be ``the shoes are quite wide''. If this evidence is used to support both arguments with the claim ``The \textit{fit} is good'' and the claim ``The \textit{fit} is bad''; then for the aspect \textit{fit}, ``the shoes are quite wide'' is a piece of controversial evidence, thus we should not include it into the final summary. 
%

Based on Definitions~\ref{def:support} and~\ref{def:contr}, we measure the global validity of an argument \textit{i} in a cluster \textit{c} by using Equation \ref{eq:val}.

\begin{equation}
    \label{eq:val}
Score({Arg_{i}^c}) = \smashoperator{\sum_{\substack{\forall Arg_j \in c, i \neq j\\ Arg_{i}.a = Arg_{j}.a}}} \hat s_i \times \hat s_j,
\end{equation}
where 
$\hat s_i$ and $\hat s_j$ represent the sentiment polarity of $Arg_i^c$  and $Arg_j^c$, respectively. An argument with a `good' sentiment is assigned a polarity value of $+1.0$, while an argument with a `bad' sentiment is assigned a polarity value of $-1.0$.



In {\argsc}, as a final step, we assign each evidence cluster with the highest score achieved by any argument within it. The clusters are then ranked based on their scores, and the top-N representative evidence pieces are selected to generate the final summary.








\section{Experimental Setup} \label{sec:exp}
In this section, we introduce the
datasets used in our experiments (Section \ref{sec:4.1:data}) and discuss the implementation details of {\argsc} (Section \ref{sec:impl}). Then we describe other comparison systems (Section \ref{sec:4.3:otherModel}), and
explain the automatic metrics for our evaluation (Section \ref{sec:4.4:eval_metric}). 

\subsection{Dataset}
\label{sec:4.1:data}
We conducted our experiments by using the {AmaSum} dataset~\cite{brazinskas-etal-2021-learning}, the largest abstractive opinion summarization dataset, consisting of more than 33,000 human-written summaries for Amazon products from a wide range of categories. In AmaSum dataset, each product is paired with more than 320 customer reviews and at least one reference summary. Each reference summary includes `verdict',
`pros' and `cons', but as the reference summaries are obtained from external resources, they are not grounded in product reviews. 
Similar to the work of \citet{hosking2023attributablescalableopinionsummarization}, we concatenate these three sections together to construct the final reference summary. Moreover, we follow the same setting to build the test set by sampling
50 products per domain for evaluation. Detailed statistics are listed in Table~\ref{tb:dataset}. 

\subsection{Implementation} \label{sec:impl}
In {\argsc} framework, we choose one closed-source LLM \textit{GPT-4o-mini} from
OpenAI\footnote{\url{https://platform.openai.com/docs/models/gpt-4o-mini}}and another open-source LLM \textit{Qwen2.5-7B} \cite{qwen2025qwen25technicalreport} as our backbone models. The prompt used for both models is shown in Appendix~\ref{app:prompt-asc}.

In order to implement the aspect clustering~(Section~\ref{sec:3.1:aspect}) and evidence clustering~(Section~\ref{sec:3.2:cluster}), 
we opt for the Density-based spatial clustering of applications with noise (\textit{DBSCAN}) algorithm \cite{10.5555/3001460.3001507}. DBSCAN is the most ideal clustering method for {\argsc} as it does not require a predefined number of clusters, thereby enhancing the generalizability of the framework. Based on a series of preliminary trials on the training set, we configure the clustering hyper-parameters as follows: the clustering metric is set to ``cosine'' similarity, the minimum number of sample per cluster is set to $1$, and the $\epsilon$ is set to $0.5$ and $0.21$ for aspect clustering and evidence clustering, respectively. Additionally, we select the top $8$ pieces of unique evidence to form our final summary based on our exploratory experiments.

\begin{table}[tb]
\centering
\begin{tabular}{lcc}
\toprule
\textbf{Test Domain}      & \begin{tabular}[c]{@{}l@{}} \textbf{\#Reviews} \end{tabular}& \begin{tabular}[c]{@{}l@{}} \textbf{Avg. Length} \end{tabular} \\ \midrule
\textit{Electronic}        & 568                 & 45                     \\ \midrule
\textit{Shoes}             & 381                 & 38                     \\ \midrule
\textit{Sports \& Outdoor} & 610                 & 44                     \\ \midrule
\textit{Home \& Kitchen}   & 680                 & 45                     \\ \bottomrule
\end{tabular}

\caption{The statistics of all the domains in our sampled test set. \textit{\#Reviews} represents the average number of reviews for all the products, and \textit{Avg.~Length} represents the average number of words separated by space in reviews for a particular domain.
 \label{tb:dataset}}
\end{table}

\subsection{Other Models for Comparison}
\label{sec:4.3:otherModel}
As depicted in Figure~\ref{img:argsum}, our proposed {\argsc} framework is a hybrid summarization approach that combines \textit{abstractive} methods (by benefiting from LLMs) and \textit{extractive} methods (by selecting the final set of arguments for summarization with clustering and TextRank). 
According to this, we primarily compare our framework with the previous state-of-the-art hybrid summarization model, HERCULES \cite{hosking2023attributablescalableopinionsummarization}. 
Since HERCULES is domain-specific, we use their released models for the four domains~(\textit{Electronic}, \textit{Shoes}, \textit{Sports \& Outdoor}, \textit{Home \& Kitchen}) as shown in Table \ref{tb:dataset}. We evaluate the models on these four domains separately using their default configuration settings.  

For comparison, 
we also develop an LLM-based baseline using GPT-4o-mini to evaluate the effectiveness of our {\argsc} framework. In this case, we randomly sample 50 reviews (the maximum number of reviews that would reliably fit within the context-length of gpt-4o-mini) and pass them to the model along with a simple summarization instruction\footnote{Prompt: \texttt{Summarize the following list of reviews. Keep your answer concise while capturing as many diverse points of view as possible.}}.

\begin{table*}[t]
    \centering
    \setlength{\tabcolsep}{10pt} 
    
    
    \begin{tabular}{lccccc} 
    \toprule
    \cmidrule(lr){2-6}
       \textbf{Models} 
       & \textbf{ROUGE-2}
       & \textbf{ROUGE-L}
       & \textbf{SC\(_{\text{ref}}\)}
       & \textbf{SC\(_{\text{in}}\)}
       & \textbf{Diversity}\\
    \midrule
    
    &\multicolumn{5}{c}{\textit{Electronics}} \\

    \cmidrule{2-6}
    GPT-4o-mini
       & \textbf{2.93}
       & 11.38
       & 20.80
       & 43.76
       & 0.55\\
    HERCULES       
       & 2.41
       & 12.44
       & 22.87
       & 79.79
       & 0.73\\
    {\argsc}\(_{\text{qwen2.5-7B}}\)
       & 2.80
       & 12.57
       & 23.91
       & 84.59
       & \textbf{0.81} \\
    {\argsc}\(_{\text{gpt-4o-mini}}\)
       & 2.68
       & \textbf{12.80}
       & \textbf{24.18}
       & \textbf{85.28 }
       & 0.80 \\
    \midrule
    &\multicolumn{5}{c}{\textit{Shoes}} \\
    \cmidrule{2-6}
    GPT-4o-mini
       & \textbf{3.75}
       & \textbf{13.23}
       & 21.46
       & 42.73
       & 0.47\\

    HERCULES  
       & 1.80
       & 12.06
       & 24.35
       & 84.45
       & 0.72\\
    {\argsc}\(_{\text{qwen2.5-7B}}\)
       & 2.14
       & 11.41
       & 25.30
       & 92.72
       & \textbf{0.75}\\
    {\argsc}\(_{\text{gpt-4o-mini}}\)
       & 2.01
       & 11.09
       & \textbf{27.09}
       & \textbf{95.28}
       & 0.72 \\
    
    
       
       
       
    \midrule
    &\multicolumn{5}{c}{\textit{Sports \& Outdoors}}\\
    \cmidrule{2-6}
    GPT-4o-mini 
        & \textbf{2.98}
        & 12.68
        & 20.69
        & 44.68
        & 0.47
        \\
    HERCULES       
       & 1.72
       & 11.45
       & \textbf{24.85}
       & 86.22
       & \textbf{0.86} \\
    {\argsc}\(_{\text{qwen2.5-7B}}\)
       &  2.20
       &  12.67
       & 24.79
       & 87.27
       & 0.82 
       \\
    {\argsc}\(_{\text{gpt-4o-mini}}\)
       & 2.65
       & \textbf{12.95}
       & 24.81
       & \textbf{89.15}
       & \textbf{0.86 } \\
       
    \midrule
        & \multicolumn{5}{c}{\textit{Home \& Kitchen}}   \\
        \cmidrule{2-6}
    GPT-4o-mini
      & \textbf{2.74}
      & 12.07
      & 20.62
      & 43.62
      & 0.55\\
    HERCULES 
    & 2.26
       & 11.35 
       & 23.31 
       & 83.24
       & 0.81\\
    {\argsc}\(_{\text{qwen2.5-7B}}\)
       &  2.45
       &  12.59
       &  \textbf{24.10}
       & 87.10
       & \textbf{0.87} \\

    {\argsc}\(_{\text{gpt-4o-mini}}\)
    & \textbf{2.74}
       & \textbf{12.80}
       & 23.66
       & \textbf{87.38}
       & 0.86 \\
       \midrule
    
    
    &\multicolumn{5}{c}{\textbf{Average}} \\
    \cmidrule{2-6}
       GPT-4o-mini
      & \textbf{3.10}
      & 12.34
      & 20.89
      & 44.68
      & 0.51\\
    HERCULES       
       & 2.05
       & 11.83
       & 23.85
       & 83.43
       & 0.78\\
    {\argsc}\(_{\text{qwen2.5-7B}}\)
       &  2.40
       &  12.31
       &  24.53
       &  87.92
       & 0.81\\
    {\argsc}\(_{\text{gpt-4o-mini}}\)
        &  2.52
       &  \textbf{12.41}
       &  \textbf{24.94}
       &  \textbf{89.27}
       & \textbf{0.81}\\
    
    \bottomrule
\end{tabular}
    \caption{Results for automatic evaluation on review summarization. ROUGE-2 and ROUGE-L F1 scores are computed against the reference summaries. SC\(_{\text{ref}}\) and SC\(_{\text{in}}\) indicate the consistency (measured using SummaC) of generated summaries against reference summaries and input reviews, respectively. Our proposed \textit{Diversity} measures the sentence-level diversity of the final generated summaries. Bold denotes the best score per domain.}
    \label{tab:eval_res}
\end{table*}

\begin{figure*}[t]
    \centering
    
    \begin{subfigure}[T]{0.3\textwidth}
        \centering
        \fbox{\parbox{\linewidth}{\textit{Great Peeler}. This product is a joke. Love this \textit{ice crusher}. Not too heavy, not too light. \textit{Easy to peel off}. \textit{Keeps my coffee hot for hours}. This \textit{ice crusher} works great. The lids fit snug. The plastic is very thin and flimsy. \textit{Crushes ice} very well. \textit{Love this water bottle}!
}}
        \caption{\sc{Hercules}}
        \label{fig:text-hercules}
    \end{subfigure}
    \hfill
    \begin{subfigure}[T]{0.3\textwidth}
        \centering
        \fbox{\parbox{\linewidth}{One tray shattered the first time we used it. I like the fact these have lids. Very easy to pop out the ice cubes. cubes end up being a little small. Lids don’t stay closed at all. Lids are nice to help keep the water in the trays when transferring from the sink to the freezer and for stacking while they make ice. Ice cubes are small. trays are very small, not easy to use as ice is hard to remove and there is only enuf ice per tray for one small glass. They stack great.
}}
        \caption{{\argsc}\(_{\text{qwen2.5-7B}}\)}
        \label{fig:text-qwen}
    \end{subfigure}
    \hfill
    \begin{subfigure}[T]{0.3\textwidth}
        \centering
        \fbox{\parbox{\linewidth}{Very easy to pop out the ice cubes. one tray broke. I like the fact these have lids. the lids do not stay on. the size of the cubes, they seem much smaller than a standard ice cube tray. cubes end up being a little small. Cubes could be a little larger. Each one comes with a lid so it's easy to stack. the silicone bottom makes them pop out with absolutely no effort.}}
        \caption{{\argsc}\(_{\text{gpt-4o-mini}}\)}
        \label{fig:text-gpt}
    \end{subfigure}
    \caption{Example generated summaries from HERCULES and {\argsc} with LLMs, for a randomly selected product (ice-tray).}
    \label{fig:text-examples}
\end{figure*}

\subsection{Evaluation Metrics}
\label{sec:4.4:eval_metric}
We use various automatic evaluation metrics to compare {\argsc} framework with other models, namely ROUGE-2, ROUGE-L F1 \cite{lin-2004-rouge}, SummaC \cite{10.1162/tacl_a_00453}. We also propose a new sentence diversity score to measure the sentence-level diversity of a summary.

We calculate the ROUGE-2, ROUGE-L F1 scores against the reference summaries of AmaSum dataset similar to the work of ~\citet{hosking2023attributablescalableopinionsummarization}. SummaC score~\cite{10.1162/tacl_a_00453} is a popular metric for evaluating how well a summary is entailed by the input document. It segments the input document and reviews into sentences and computes the average entailment score between each pair of the input sentence and the generated summary. We calculate the SummaC score of the generated summaries against the reference (\textit{$SC_{ref}$}) and the original input reviews (\textit{$SC_{in}$}). 
Since the reference summary is built independently of the input reviews, the SummaC score computed against original reviews (\textit{$SC_{in}$}) provides a more trustworthy indication of the summary quality.

A helpful product review summary should capture the most frequently expressed opinions from the input, but without repeating the same points redundantly. Therefore, we propose a diversity metric that evaluates the sentence-level diversity of the final summary. The idea is to segment a summary into sentences and evaluate the semantic closeness of all the sentences through clustering. As a longer summary having more sentences would result in a higher number of clusters naturally, we normalise the cluster number by the total number of sentences to obtain the final diversity score of a summary. We define this new metric in Equation~\ref{eq:diversity}. 

\begin{equation}
\label{eq:diversity}
    Diversity(S) = \frac{|Clusters(S)|}{|S|},
\end{equation}
where $S$ is the set of sentences in a summary, $|Clusters(S)|$ is the number of clusters and $|S|$ is the number of sentences in S.

In our implementation, we use DBSCAN algorithm with the same parameter settings as the aspect clustering discussed in Section~\ref{sec:impl}.

    

%

\section{Evaluation Results} \label{sec:eval}
In this section, we analyze the quantitative results based on all automatic evaluation metrics (Section~\ref{sec:eval:numeric}) and provide a detailed qualitative discussion on the generated summaries for a randomly chosen product (Section~\ref{sec:eval:quality}).

\subsection{Quantitative Analysis}
\label{sec:eval:numeric}
The evaluation results are shown in Table \ref{tab:eval_res}. We observe that {\argsc} framework with both closed-source and open-source LLMs consistently outperforms other methods on all four domains across all metrics besides ROUGE-2. Particularly for the SC\(_{\text{in}}\) score, our 
{\argsc} achieves significantly higher SC\(_{\text{in}}\) scores across all the domains, indicating that our generated summaries are more representative of the input reviews. Surprisingly, our {\argsc} framework paired with Qwen2.5-7B~({\argscqwen}) achieves comparable performance with {\argsc} paired with GPT-4o-mini~({\argscmini}) across all the domains and evaluation metrics, demonstrating both the robustness and the generalizability of the framework. 

Across all models, the big difference between the SC\(_{\text{in}}\) and SC\(_{\text{ref}}\) score also suggests that the manually constructed reference summaries do not faithfully entail all the product reviews, as they are built separately. On the other hand, GPT-4o-mini baseline performs the worst on most of the metrics, which can be the result of the limited number of input reviews. However, it achieves higher ROUGE scores and has a smaller difference in SC\(_{\text{ref}}\) than it has in SC\(_{\text{in}}\) when compared to other methods. This indicates that summaries generated by GPT-4o-mini are more fluent and closer to manually written summaries.


In terms of the sentence-level diversity, {\argscqwen} even performs better than {\argscmini} in most domains. 
Notably, the diversity of summaries generated by our {\argsc} framework is greatly dependent on the diversity of unique aspects of products. For domains having products with various aspects, such as \textit{Electronics} (on average 14 aspects per product ), the diversity score of our summaries is obviously higher than other domains, such as \textit{Shoes} (on average 10 aspects per product).
While {\argsc} with LLMs generate less diverse summaries for the \textit{Shoes} and \textit{Sports \& Ourdoors} domains, it achieves higher SummaC scores on these domains compared to the others. 
This reveals that a summary could attain a high SummaC score by repeating opinions closely aligned with the input documents, even if such a summary may not be considered helpful in a real-life setting.


\subsection{Qualitative Analysis}
\label{sec:eval:quality}
In addition to the numerical results in Table \ref{tab:eval_res}, we randomly select one example product from the \textit{Home \& Kitchen} domain to discuss qualitative aspects of our generated summaries. As shown in Figure \ref{fig:text-examples}, we notice that our generated results are significantly more faithful to the original reviews. This is because {\sc{HERCULES}} decodes the summary from a hierarchical discrete latent embedding space, 
which strongly relies on its pre-trained codebook that performs the mapping from the discrete code to continuous embeddings \cite{hosking2023attributablescalableopinionsummarization}. However, since the codebook is pre-trained on the training set, for an uncommon product in the training set, their model would struggle to encode the reviews properly and decode the relevant information accordingly. This is also justified by the unsatisfying performance of {\sc{HERCULES}} in the \textit{Electronics} domain, where the types of products are more diverse than in other domains.
On the contrary, our summaries maintain the topic at hand and minimise the likelihood of hallucination as we only apply abstractive summarization in the initial aspect-centric argument extraction step. 
In addition, by comparing the textual summary from {\argscqwen} and {\argscmini}, we observe that evidence extracted by GPT-4o-mini is summarized to be more concise, which may lead to a lower diversity score for some domains.

\section{Conclusion} \label{sec:conc}

This paper presents a novel summarization framework that integrates aspect-based sentiment analysis with argument mining to extract aspect-centric arguments for
generating diverse yet faithful summaries. Although evaluating arguments based on their controversy level may not be the most ideal solution, our approach 
obtains strong performance on a benchmark dataset in both numerical and qualitative evaluations. Furthermore, by combining both extractive and abstractive summarization techniques, we demonstrate strong generalisability of our framework through automated aspect generation, the incorporation of multiple LLMs and domain-independent summarization. 

Our approach relies on the dynamic extraction of relevant aspects and sentiments towards these aspects. We are planning to use these aspects to generate summaries as part of our future work.
We will also conduct user studies to find meaningful ways to present the summary together with this aspect sentiment structure. Future research should also focus on finding ways to automatically evaluate structured summaries, which remains as a challenging problem for the community.

\section*{Limitations}
{\argsc} framework can be easily adapted to other domains and incorporated with other language models; however, we have a number of hyper-parameters set to run the clustering algorithm. 
The consistent performance of our framework across four domains suggests the generalisability of this set of chosen parameters, but it may require more adjustments when adapting to new datasets. 
Besides, since our summaries are generated by concatenating pieces of evidence from different arguments, they may lack coherence in general.

\section*{Acknowledgment}
This work was supported by the University of Edinburgh-Huawei Joint Lab grants CIENG4721 and CIENG8329.


\bibliography{custom}
\newpage
\appendix

\section{Appendix A}
\label{app:prompt-asc}

We provide the prompt used in our paper in Table~\ref{tab:prompt-argsc}.

\begin{table}[htb]
    \centering
    \footnotesize
    \begin{tabularx}{0.48\textwidth}{|>{\raggedright\arraybackslash}X|}
         \hline
Fill the scheme with the provided review. 

\{\textbf{{Review Argumentation Scheme}}\}

Note:\\
1. Identify the aspects mentioned in the review. 
Then provide a new scheme with the relevant evidence for each identified aspect. \\
2. The most mentioned aspects are \{\textbf{aspect}\}. \\
3. Only generate a new aspect when there is no matching one above.\\
4. Do NOT provide scheme having aspect wasn't mentioned in the text.\\
5. Do NOT include too much details in the evidence.\\

Please return the values in JSON format: \\

[\{``aspect'': ``the property / feature of the product'', \\``sentiment'': ``positive/negative'', \\``evidence'': ``support from the argument''\}, ...]\\

\hline
    \end{tabularx}
    \caption{Prompt provided to {\argscmini} and {\argscqwen}, where ``{Review Argumentation Scheme}'' is the placeholder to fit in the RAS (Table \ref{tab:review_scheme}) and the ``{aspect}'' is the placeholder to interactively input the most popular aspects we have in the current aspect set.}
    \label{tab:prompt-argsc}
\end{table}








\end{document}